\pdfoutput=1

\documentclass[11pt]{article}

\usepackage[]{ACL2023}

\usepackage{times}
\usepackage{latexsym}
\usepackage{multirow}
\usepackage{amsfonts, amsmath, amssymb,mathtools,bbm,bm}
\usepackage[T1]{fontenc}

\usepackage[utf8]{inputenc}

\usepackage{microtype}

\usepackage{inconsolata}
\usepackage{amsthm}
\usepackage{makecell}
\usepackage[frozencache,cachedir=minted-cache]{minted}

\theoremstyle{plain}

\title{Dissecting Transformer Length Extrapolation via \\ The Lens of Receptive Field Analysis}

\author{Ta-Chung Chi \\
  Carnegie Mellon University \\
  \texttt{tachungc@andrew.cmu.edu} \\\And
  Ting-Han Fan \\
  Princeton University \\
  \texttt{tinghanf@princeton.edu} \\\AND
  Alexander I. Rudnicky \\
  Carnegie Mellon University \\
  \texttt{air@cs.cmu.edu} \\\And
  Peter J. Ramadge \\
  Princeton University \\
  \texttt{ramadge@princeton.edu}
}

\begin{document}
\maketitle
\begin{abstract}
Length extrapolation permits training a transformer language model on short sequences that preserves perplexities when tested on substantially longer sequences.
A relative positional embedding design, ALiBi, has had the widest usage to date. 
We dissect ALiBi via the lens of receptive field analysis empowered by a novel cumulative normalized gradient tool. The concept of receptive field further allows us to modify the vanilla Sinusoidal positional embedding to create  ~\textbf{Sandwich}, the first parameter-free relative positional embedding design that truly length information uses longer than the training sequence. Sandwich shares with KERPLE and T5 the same logarithmic decaying temporal bias pattern with learnable relative positional embeddings; these elucidate future extrapolatable positional embedding design.
\end{abstract}

\section{Introduction}
The length of input sequences is an important hyperparameter choice for pretraining a transformer language model. 
A vanilla transformer language model has a quadratic training cost w.r.t $L_{tr}$, the training sequence length. 
As  the value of $L_{tr}$ increases, cost becomes impractical. 
However, we can use the model for substantially longer evaluation sequence lengths $L_{ex}\gg L_{tr}$ as gradients no longer need to be recorded. 
The discrepancy between $L_{tr}$ and $L_{ex}$  motivates the task of~\textbf{length extrapolation}~\cite{press2022train}:~\emph{Can a transformer language model maintain equally good, if not better, perplexities when longer sequences are used in the testing stage?}

Several extrapolatable transformer language models have been proposed including ALiBi~\cite{press2022train} and KERPLE~\cite{chi2022kerple}, of which the relative positional embedding design is hypothesized to be critical to success.
Empirically, they extrapolate to $L_{ex} \gg L_{tr}$ much better than other absolute and relative positional embeddings including Sinusoidal~\cite{vaswani2017attention}, Rotary~\cite{su2021roformer}, and T5~\cite{raffel2019exploring}, resulting in the adoption of ALiBi for the recently released Bloom~\cite{scao2022bloom} model.
\begin{figure}
    \centering
    \includegraphics[width=\linewidth]{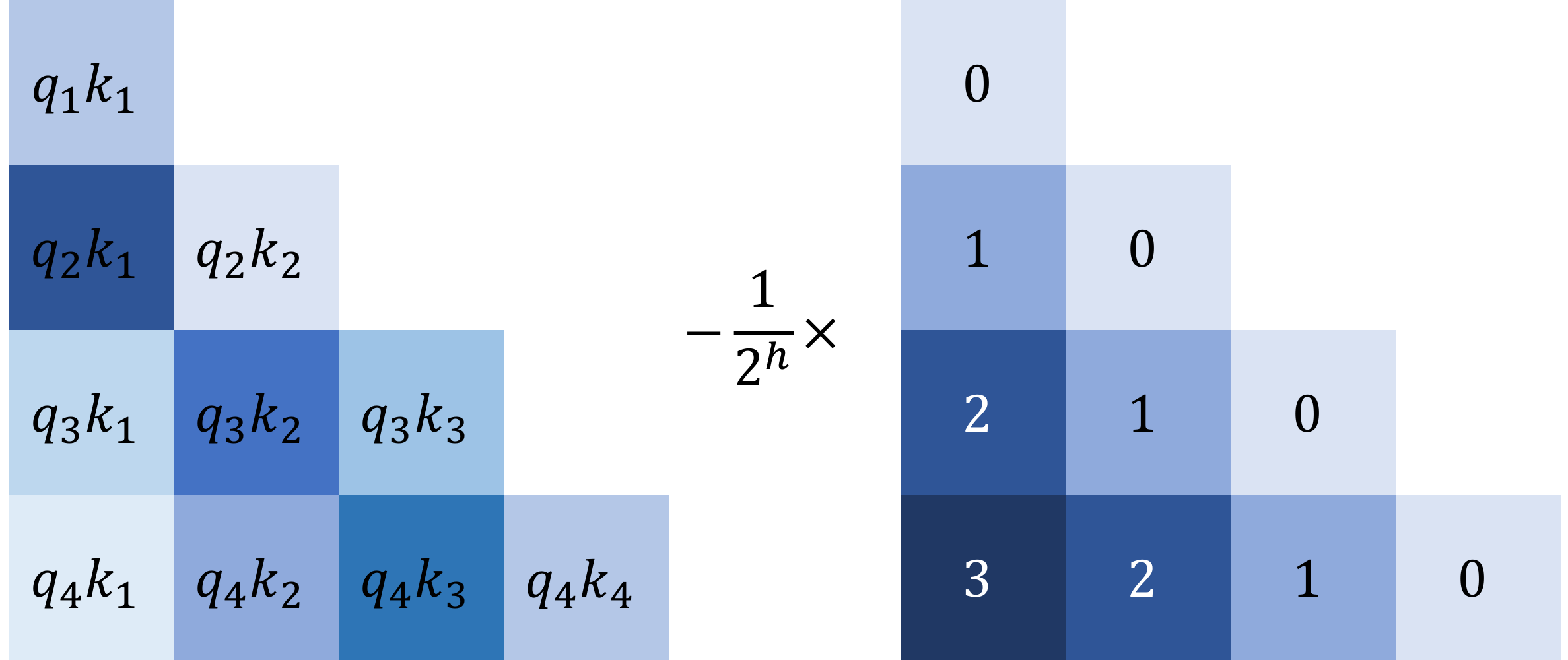}
    \caption{ALiBi. For a transformer language model with $H$ attention heads, the range of $h$ is $n\cdot\frac{8}{H}$, where $n=\{1\dots H\}$. Left = self-attention matrix, right = temporal biases matrix.}
    \label{fig:alibi}
\end{figure}
\begin{figure}
    \centering
    \includegraphics[width=\linewidth]{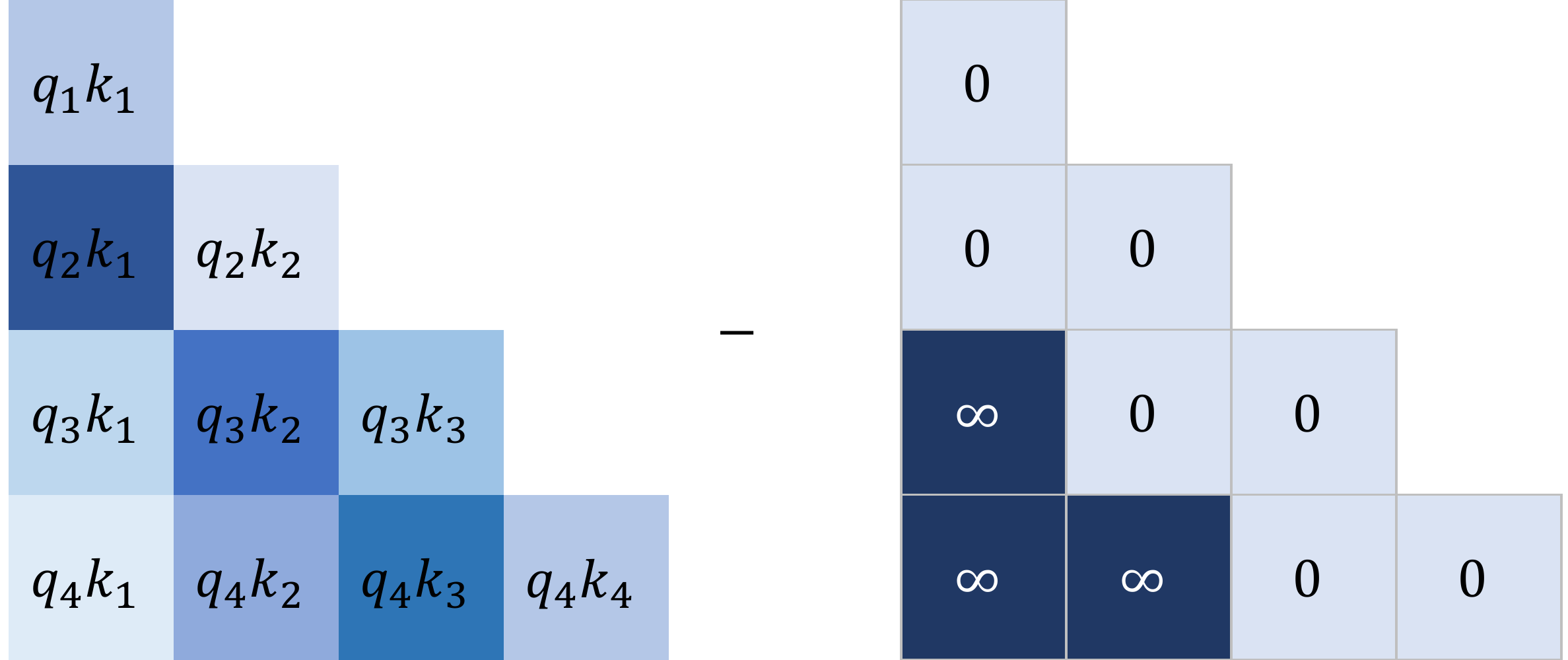}
    \caption{Windowed Attention. This is the same design as Longformer~\cite{beltagy2020longformer}. We limit the context window size to $w=2$ in this example. Left = self-attention matrix, right = temporal biases matrix.}
    \label{fig:window}
\end{figure}
Despite the significant empirical success of ALiBi, there is still a lack of fundamental understanding of why it works.\footnote{\url{https://github.com/ofirpress/attention_with_linear_biases\#why-do-you-think-alibi-works}}

Figure~\ref{fig:alibi} shows the implementation of ALiBi. We hereinafter refer to the coefficient $\frac{1}{2^h}$ as~\emph{slope}. 
Intuitively, ALiBi encourages a token to focus on neighbors based on its temporal biases matrix. 
When two tokens are distant, ALiBi becomes highly similar to windowed attention, shown in Figure~\ref{fig:window}.
Experiments in \S\ref{sec:three_exps} will further establish the connection between the two.

Windowed attention allows the easy derivation of a theoretical (maximum) receptive field: $wR$ for an $R$ layer transformer model with windowed attention size $w$. 
A windowed attention model can extrapolate if $L_{tr}>wR$ because 1) $wR$ is fully covered by $L_{tr}$ during the training stage, and 2) it simply ignores the additional $L_{ex}-wR$ tokens during the testing stage. 
Surprisingly, a model can still extrapolate when $L_{tr}<wR$ which we show in \S\ref{sec:three_exps}. 
This calls for the need for empirical receptive field measurement and motivates our model-agnostic cumulative normalized gradient tool. The tool we develop can be applied back on ALiBi to show that $L_{tr}$ covers most of its empirical receptive field.

Our analysis tool also provides critical context for explaining the length extrapolation failure~\cite{press2022train,chi2022kerple} of Sinusoidal~\cite{vaswani2017attention} and Rotary~\cite{su2021roformer} by showing their violation of the empirical receptive field coverage principle.
Sinusoidal can be fixed by dropping the intermediate terms and keeping only the decay-with-distance biases; this leads to the creation of~\textbf{Sandwich}, the first parameter-free relative positional embedding that uses information beyond  $L_{tr}$. Sandwich shares a similar temporal bias pattern with trainable positional embeddings such as KERPLE~\cite{chi2022kerple} and T5~\cite{raffel2019exploring}, and they jointly suggest the future design of extrapolatable transformer positional embeddings.

\section{Related Work}
\subsection{Length Extrapolation}
\label{sec:related_le}
In the context of language modeling, we expect token-level perplexities to remain at least the same, if not lower (i.e. better), when $L_{ex}\gg L_{tr}$ sequences are provided. 
Recurrent neural networks~\cite{MikolovKBCK10,6424228,zaremba2014recurrent} can easily perform length extrapolation. 
But this is not an easy task for transformer language models, among which only those equipped with special relative positional embeddings~\cite{press2022train, chi2022kerple} are length extrapolatable.

\subsection{Positional Embeddings}
It is widely believed that the design of positional embeddings is the key to successful length extrapolation of transformer language models~\cite{press2022train,chi2022kerple}. We can roughly categorize existing positional embeddings into absolute (APE)~\cite{vaswani2017attention} and relative (RPE)~\cite{su2021roformer,raffel2019exploring,press2022train,chi2022kerple} variants. 
APE often assigns one positional embedding per token and combines them directly with input embeddings. 
In contrast, RPE adds temporal bias terms to the self-attention matrix to encode the relative distance between token pairs. For example, the right triangular matrix in Figure~\ref{fig:alibi} shows the set of temporal bias terms.
It is challenging for APE to extrapolate well without any further fine-tuning since either the beyond $L$ positional embeddings do not exist, or the model needs to process unseen positional embeddings (e.g. unseen sinusoidal embeddings).~\cite{press2022train,chi2022kerple}. In contrast, RPE usually performs better length extrapolation since it is easier to construct the additional temporal bias terms.

\subsection{Windowed and Sparse Attention}
We will see later that ALiBi can be viewed as imposing a windowed attention mask on the self-attention matrix, similar to previous transformer models with sparse attention~\cite{beltagy2020longformer,bigbird,etc,gmat}. 
Interpreting ALiBi from the perspective of windowed attention allows us to easily calculate the theoretical receptive field of a model.

\subsection{Receptive Field}
A model's receptive field is defined as the size of the input region that contributes the most to model outputs. It is often measured in the context of convolution neural networks~\cite{receptive, deformable,araujo2019computing,raghu2021do,dosovitskiy2021an} and their dilated variants~\cite{oord2016wavenet, dilated_conv,dilated_rnn, beltagy2020longformer} with the ultimate goal of receptive field size maximization. 
Even though we focus on transformer language models, we borrow the idea to show that the empirical receptive field coverage of a model is crucial to its length extrapolation performance.

\section{Background and Notations}
\subsection{Transformer Language Model}
Given a sequence of $L\in\{L_{tr}, L_{ex}\}$ input embeddings $\{\bm e_m\}_{m=1}^L$ in $\mathbb{R}^d$, an $R$ layer transformer language model with $H$ attention heads converts each $\bm e_m$ into its corresponding query, key, and value vectors in $\mathbb{R}^\frac{d}{H}$ at each layer:
\begin{align*}
    \bm q_m=\bm W_q\bm e_m,\enskip\bm k_m=\bm W_k\bm e_m,\enskip\bm v_m=\bm W_v\bm e_m,
\end{align*}
where $\bm W_q$, $\bm W_k$, $\bm W_v\in\mathbb{R}^{\frac{d}{H} \times d}$ are learnable matrices. The resulting vectors are processed by the self-attention module for pre-Softmax logits:
\begin{equation*}
    l_{mn}= 
\begin{cases}
    \langle \bm q_m, \bm k_n \rangle,& \text{if } m\geq n\\
    -\inf,              & \text{otherwise}
\end{cases}
\end{equation*}
followed by the scaled softmax normalization:
\begin{equation}
\label{eq:scaled_softmax}
    a_{m,n}=\frac{\exp(l_{m,n}/\sqrt{d/H})}{\sum_{i=1}^L \exp(l_{m,i}/\sqrt{d/H})}
\end{equation}
To be precise, the matrices ($\bm W_q^{(h)}$, $\bm W_k^{(h)}$, $\bm W_v^{(h)}$), vectors ($\bm q_m^{(h)}$, $\bm k_m^{(h)}$, $\bm v_m^{(h)}$, $\bm o_m^{(h)}$), and scalars ($l_{mn}^{(h)}$, $a_{mn}^{(h)}$) are associated with a head number $h$. For notation simplicity, we only show the dependency on $h$ when we need it. For example, the output vector $\bm o_m^{(h)}$ at position $m$ for head $h$ is:
\begin{equation*}
    \bm o_m^{(h)} = \sum_{n=1}^L a_{m,n}^{(h)}\bm v_n^{(h)}
\end{equation*}
All the $H$ output vectors are concatenated, denoted by $\oplus$, and transformed by $\bm W_o\in\mathbb{R}^{d\times d}$ to obtain $\bm o_m\in \mathbb{R}^{d}$:
\begin{equation*}
    \bm o_m = \bm W_o (o_m^{(1)} \oplus o_m^{(2)} \oplus \cdots \oplus o_m^{(H)})
\label{eq:om}
\end{equation*}
A layer normalization \cite{ba2016layernorm} on $\bm o_m$, i.e. $\text{LayerNorm}(\bm o_m)$, gives the input embedding to the next layer. After $R$ layers of propagation, the last $\bm o_m$ is transformed by $\bm V\in \mathbb{R}^{v\times d}$ and normalized by Softmax to get the distribution $\bm p\in \mathbb{R}^v$ over vocabulary size $v$:
\begin{equation}
    \bm p = \text{Softmax}(\bm V\bm o_m)
    \label{eq:prob}
\end{equation}
We set $R=12$, $H=12$, $d=768$, and $L_{tr}=512$ for all experiments reported in this paper.
\subsection{ALiBi}
ALiBi modifies $l_{m,n}$ to be:
\begin{equation}
    l_{mn}= 
\begin{cases}
    \langle \bm q_m, \bm k_n \rangle - \frac{1}{2^h}(m - n) ,& \text{if } m\geq n\\
    -\inf,              & \text{otherwise}
\end{cases}
\label{eq:alibi}
\end{equation}
The range of $h$ is $n\cdot\frac{8}{H}$, where $n=\{1\dots H\}$.
\subsection{Windowed Attention}
If the windowed attention has a size $w$, then:
\begin{equation*}
    l_{mn}= 
\begin{cases}
    \langle \bm q_m, \bm k_n \rangle,& \text{if } n+w> m\geq n\\
    -\inf,              & \text{otherwise}
\end{cases}
\end{equation*}

\begin{figure}
    \centering
    \includegraphics[width=\linewidth]{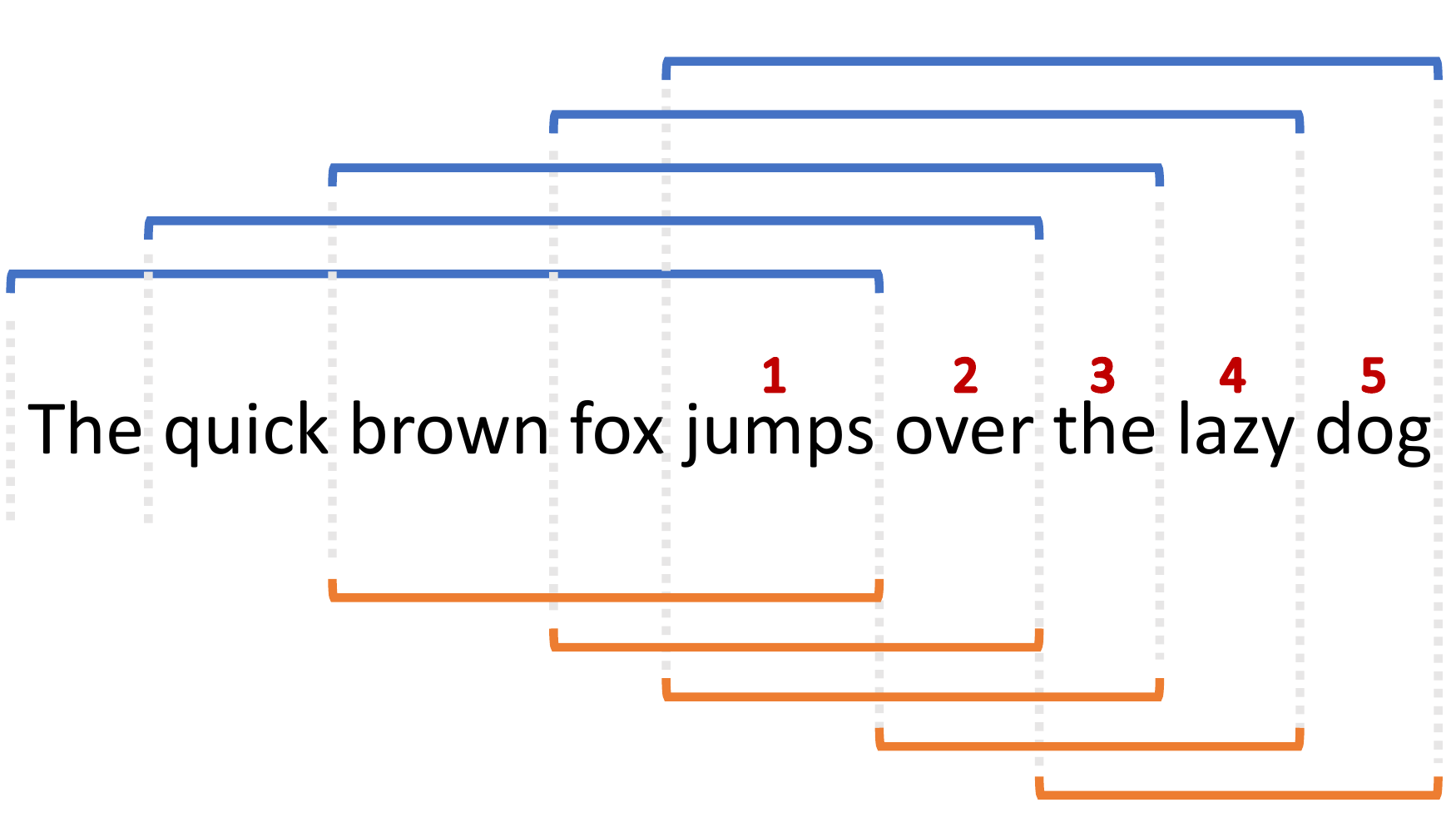}
    \caption{We always evaluate the perplexities of the 5 tokens numbered from 1 to 5. The upper brackets represent $ L_{ex}=5$. The lower brackets represent $L_{ex}=3$. This formulation ensures the same 5 tokens are always evaluated with different numbers of previous tokens.}
    \label{fig:evaluation}
\end{figure}

\subsection{Evaluation of Length Extrapolation}
We prepare $N=1000$ text segments of length $L_{ex}>L_{tr}$ from the evaluation dataset. For each segment, we alter the number of previous tokens ranging from 1 to $L_{ex} - 1$ of the last token and only calculate its perplexity:
\begin{equation*}
    \text{PPL}=\exp\left(\frac{1}{N}\sum_{i=1}^N -\log p_i\right),
\end{equation*}
where $p_i$ is the predicted probability from Eq.~(\ref{eq:prob}) of the last ($L_{ex}$-th) token in the $i$-th segment.
This ensures that the same set of tokens is always used for perplexity calculation and only their number of previous tokens is varied,  see Figure~\ref{fig:evaluation}.\footnote{There exists another evaluation protocol named non-overlapping subsequences adopted in the main experiment tables of ALiBi~\cite{press2022train}. It is not the most suitable protocol for length extrapolation evaluation as it suffers from the ``early token'' curse. Please refer to Appendix B of ALiBi~\cite{press2022train} for details.}

\begin{table*}[!ht]
    \setlength{\tabcolsep}{2pt}
    \hspace{-7mm}
    \begin{tabular}{@{\extracolsep{3pt}}lccccccccccccccccccccc}
    \hline\hline
    \multirow{2}{*}{$L_{ex}$} & \multicolumn{6}{c}{Shift all $h$ by $\Delta$ } & & \multicolumn{5}{c}{Same $h$ for all heads} & & \multicolumn{6}{c}{Windowed Attention with Size $w$} \\
     \cline{2-7} \cline{9-13} \cline{15-20} &
     $\Delta$:-3 & 0 & 2 & 4 & 6 & 8 & & $h$:0 & 2 & 4 & 6 & 8 & & $w$:40 & 80 & 100 & 120 & 160 & 320 \\
    \hline
    512 & 5.76 & 5.57 & 5.50 & 5.63 & 5.70 & 5.70 & & 9.45 & 6.65 & 5.85 & 5.60 & 5.70 & & 8.27 & 7.28 & 7.04 & 6.77 & 6.41 & 6.04 \\
    1024 & 7.15 & 5.64 & 5.31 & 5.81 & 55.4 & 55.4 & & 9.20 & 7.01 & 8.66 & 25.4 & 55.4 & & 8.27 & 7.29 & 7.02 & 8.90 & 67.4 & 178 \\
    2048 & 7.15 & 5.94 & 5.89 & 6.92 & 94.4 & 94.4 & & 9.21 & 7.08 & 8.66 & 31.7 & 94.4 & & 8.27 & 7.29 & 7.03 & 8.90 & 67.5 & 202 \\
    4096 & 7.15 & 5.95 & 5.92 & 6.94 & 96.0 & 96.0 & & 9.21 & 7.08 & 8.66 & 31.8 & 96.0 & & 8.27 & 7.29 & 7.02 & 8.90 & 67.5 & 202 \\
    8192 & 7.15 & 5.95 & 5.92 & 6.94 & 96.0 & 96.0 & & 9.21 & 7.08 & 8.66 & 31.8 & 96.0 & & 8.27 & 7.29 & 7.02 & 8.90 & 67.5 & 202 \\
    \hline\hline
    \end{tabular}
    \caption{The three experiments on the Arxiv dataset.}
    \label{tab:arxiv}
\end{table*}

\section{ALiBi and Windowed Attention}
\label{sec:three_exps}
Here, we alter the slope ($\frac{1}{2^h}$) of ALiBi to check if the length extrapolation property persists and reveal the connection between ALiBi and windowed attention. 
We present three experiments on two datasets, ArXiv and OpenWebText2 (Appendix~\ref{sec:appendix}), to ensure that the observations are consistent across different text domains, shown in Table~\ref{tab:arxiv} and~\ref{tab:openwebtext2}.
 
\subsection{Slope Shift (Shift all $h$ by $\Delta$)} 
We first investigated whether slope diversity (each attention head has one slope) is the key to length extrapolation. We shift $h$ by a fixed amount $\Delta$ and find that the model, unfortunately, fails to extrapolate beyond a certain quantity. 
This implies that diversity itself might not be the deciding factor, but that the actual slope value is more important.

\subsection{Slope Equalization (Same $h$ for all heads)}
To identify the slope magnitude that enables length extrapolation, we set all slopes to be the same instead of the original geometric sequence. We then steadily increase the slope value from 0 to 8 and find that only large slopes ($\frac{1}{2^h}$), or equivalently small $h$, allow a model to extrapolate well. Large slopes implicitly enforce a narrow windowed bias on the self-attention matrix such that distant tokens cannot interact with each other.

\subsection{Windowed Attention (Size $w$)} 
We make the implicit window effect explicit as shown by Eq.~(\ref{eq:alibi}), which is also adopted by Longformer~\cite{beltagy2020longformer}. We define the windowed attention size to be $w$. The model underperforms at small $w$ and diverges on long $L_{ex}$ at large $w$. The same trend holds in the first two experiments when $h$ is too small or large.

\subsection{Other Observations}
First, ALiBi does not in fact extrapolate since its perplexities all increase instead of staying the same when $L_{ex}>L_{tr}$. In contrast, windowed attention models are extrapolatable up to $w=100$.
Second, we can clearly see that once $L_{ex}$ passes a certain threshold, the perplexity either remains the same or explodes. This suggests that the model is either ignoring tokens beyond a certain length (same)\footnote{A limited but similar observation was made in Appendix B.2 of ALiBi~\cite{press2022train}.} or not using it properly (explosion).
In the next section, we will use the concept of receptive field to explain these observations.

\begin{figure*}[!htb]
\minipage{0.5\textwidth}
  \includegraphics[width=\linewidth]{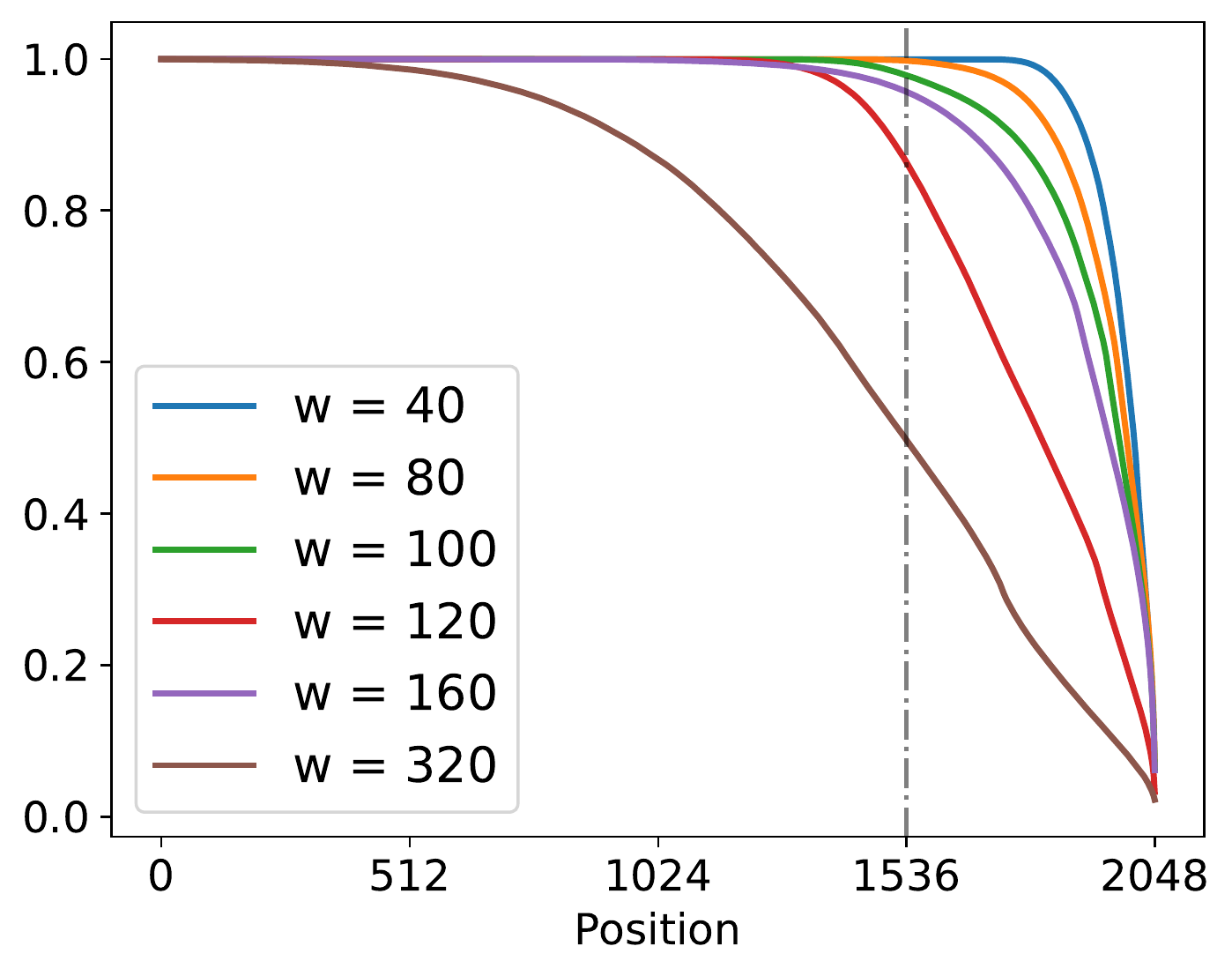}
  \caption{Cumulative normalized gradient 
 on ArXiv \\ when predicting the next (2048-th) token.}\label{fig:arxiv_sec3}
\endminipage\hfill
\minipage{0.5\textwidth}
  \includegraphics[width=\linewidth]{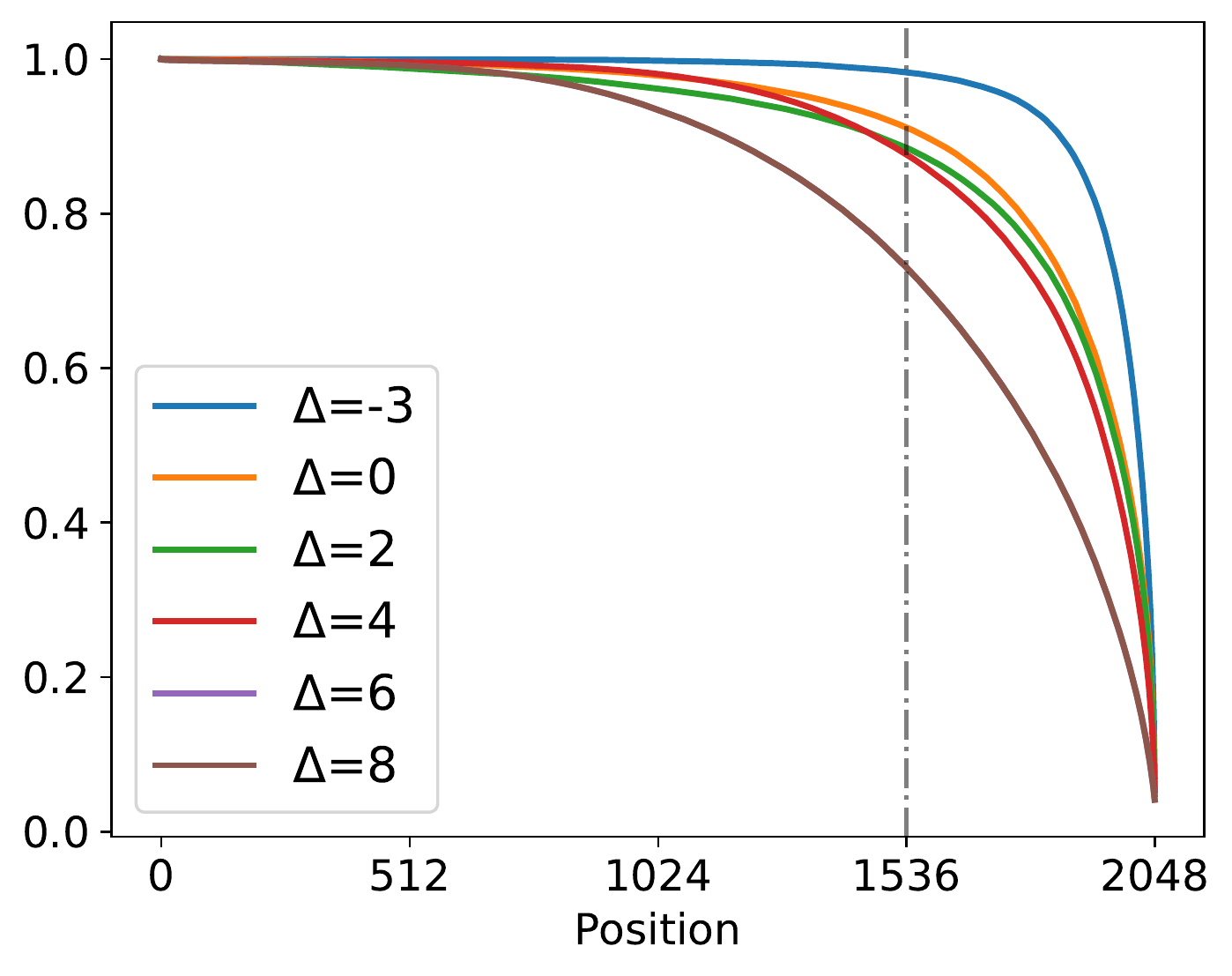}
  \caption{Cumulative normalized gradient on ArXiv \\ when predicting the next (2048-th) token.}\label{fig:arxiv_sec1}
\endminipage
\end{figure*}

\section{Receptive Field Measurement}
\label{sec:receptive}
Following the definition of windowed attention size $w$, an $R$ layer transformer has a theoretical receptive field (TRF) of $wR$, which is the maximum number of tokens that contribute to the prediction of the next token. In practice, a neural model often uses a subset of TRF, named empirical receptive field (ERF). While previous work~\cite{receptive, deformable,araujo2019computing,raghu2021do,dosovitskiy2021an,beltagy2020longformer} aims to increase ERF to match TRF, we show that decreasing ERF could serve as one feasible approach to enable successful length extrapolation.

Consider the case where TRF $\leq L_{tr}$: This model can extrapolate easily because its TRF is fully covered and trained. Concretely, if we set $R=12$, $L_{tr}=512$ in Table~\ref{tab:arxiv} and~\ref{tab:openwebtext2}, we know that as long as $w<42.6=512/12$, TRF will be fully covered by $L_{tr}$. 
Surprisingly, the model is still able to extrapolate up to $w=100$, leading to a TRF of $100*12=1200 \gg 512$. This can be explained by the ERF and TRF discrepancy discussed above; this calls for the need to quantify ERF.

\subsection{Quantifying Empirical Receptive Field}
\label{sec:quant_cm}
We first calculate the normalized gradient~\cite{receptive} of each input token w.r.t the prediction of the next token:
\begin{equation*}
    s_m=\frac{\|\bm g_m\|_2}{\sum_{n=1}^{L_{ex}} \|\bm g_n\|_2},
\end{equation*}
where $\bm g_m$ is the gradient vector of the input embedding $\bm e_m$.
We then calculate the cumulative sum as:
\begin{equation*}
    c_m=\sum_{n=m}^{L_{ex}} s_n,\quad 0 \leq c_m \leq 1,
\end{equation*}
Visualizations of $c_m$ for the slope shift and windowed attention experiments are shown in Figures~\ref{fig:arxiv_sec3} and~\ref{fig:arxiv_sec1}.
We define the ERF of a model as: $$\text{ERF} = \min \{m \mid c_m > 0.99\}.$$ Figure~\ref{fig:arxiv_sec3} demonstrates how we derive the model's ERF when it is predicting the 2048-th token. For models with $w\in [40, 80, 100]$, the most recent $L_{ex}=L_{tr}=512$ (1536-th to 2047-th) covers more than 99\% of the total (1.0) normalized gradient, so their ERF is smaller than 512. In contrast, models with $w\in [120, 160, 320]$ have ERF = 768, 1024, and 1536 tokens, respectively. Since $L_{tr}=512$ does not fully cover their ERFs, they fail to extrapolate well.

We next focus on the more complex Figure~\ref{fig:arxiv_sec1}, in which neither of the configurations reaches 0.99 within the most recent $L_{tr}=512$ tokens. Generally, this explains why the perplexity often bumps up when $L_{ex}$ goes from 512 to 1024: Models cannot perfectly process more tokens than they were trained on. If we take a closer look, the $\Delta=-3$ model has the strongest windowing effect and the smallest ERF=768 tokens, therefore its perplexity plateaus the soonest at $L_{ex}=1024$ in Table~\ref{tab:arxiv}. The remaining models all need ERF=2048 tokens to reach $c_m=0.99$, which explains why their perplexities become stable only after $L_{ex}=2048$ (Table~\ref{tab:arxiv}). For $\Delta\in[6,8]$ models specifically, the difference between $L_{tr}$ and ERF is too large to be handled, resulting in exploded perplexities.

\subsection{Fixing Failed Cases}
We fix the failed cases in Table~\ref{tab:arxiv} section 1 (varying $\Delta$) and section 3 (varying $w$) by increasing $L_{tr}$ to cover their ERFs. We increase $L_{tr}$ to 1024 for windowed attention with $w=160$; For shifted ALiBi with $\Delta=6$, we need $L_{tr}=2048$ tokens. Table~\ref{tab:extended_training_length} shows that both are now able to maintain stable perplexities.

\begin{table}[!htbp]
    \setlength{\tabcolsep}{2pt}
    \resizebox{\columnwidth}{!}{%
    \begin{tabular}{@{}lccccc}
    \hline\hline
    \multirow{2}{*}{$L_{ex}$} & \multicolumn{2}{c}{Shift all $h$ by $\Delta=6$} & & \multicolumn{2}{c}{\makecell{Windowed Attention\\ $w=160$}} \\
     \cline{2-3} \cline{5-6} &
     Arxiv & OpenWebText2 & & Arxiv & OpenWebText2 \\
    \hline
    2048 & 4.4 & 15.2 & & 6.2 & 19.9\\
    4096 & 6.2 & 19.8 & & 6.2 & 19.9\\
    8192 & 6.2 & 19.9 & & 6.2 & 19.9\\
    \hline\hline
    \end{tabular}
    }
    \caption{Fixing failed cases with longer $L_{tr}$: $L_{tr}=2048$ for ALiBi with $\Delta=6$ and $L_{tr}=1024$ for windowed attention with $w=160$.}
    \label{tab:extended_training_length}
\end{table}

\begin{figure*}[!htb]
\minipage{0.5\textwidth}
  \includegraphics[width=\linewidth]{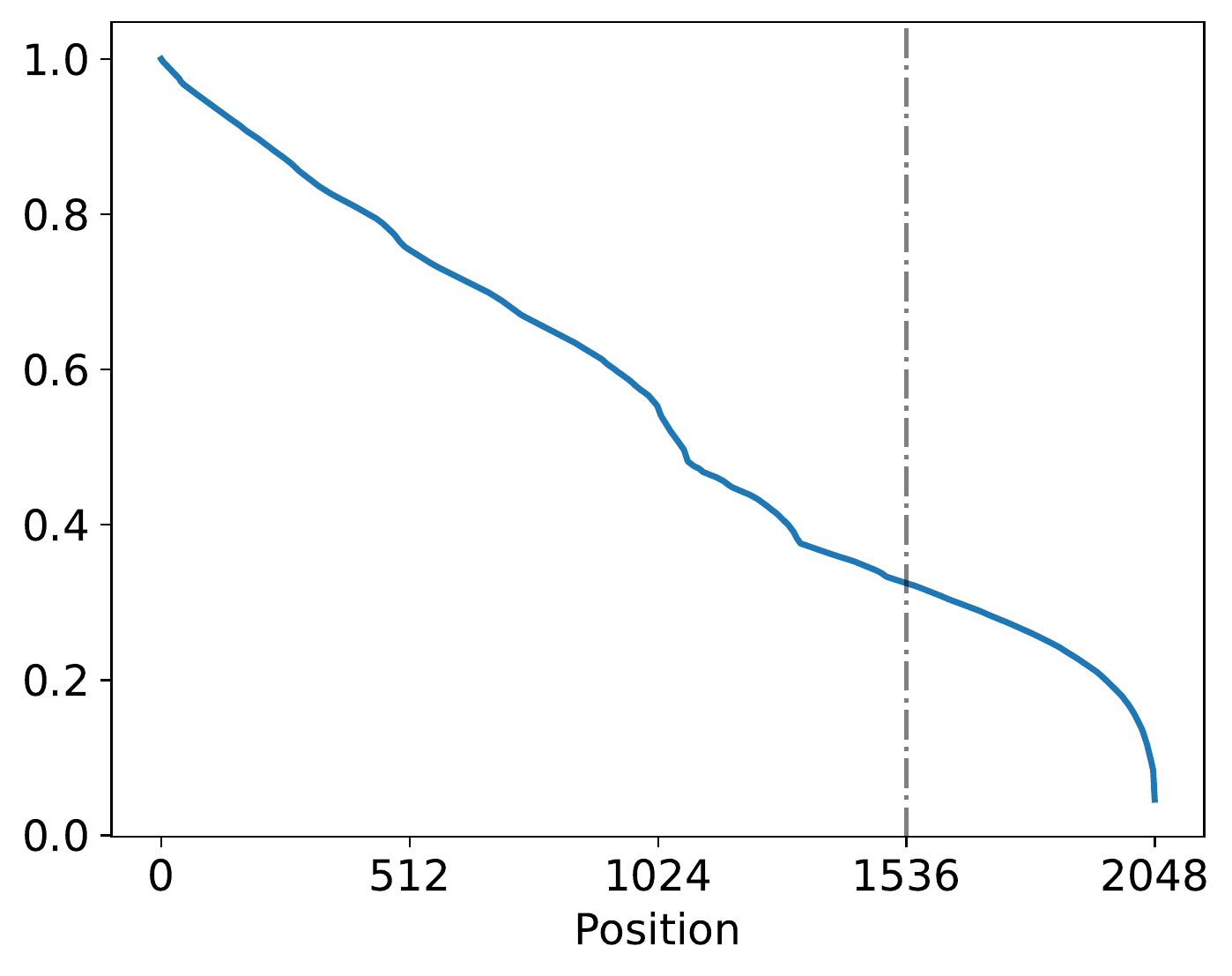}
  \caption{Cumulative normalized gradient of Rotary \\ on ArXiv when predicting the last (2048-th) token \\with $L_{tr}=512$.}\label{fig:rotary_512}
\endminipage\hfill
\minipage{0.5\textwidth}
  \includegraphics[width=\linewidth]{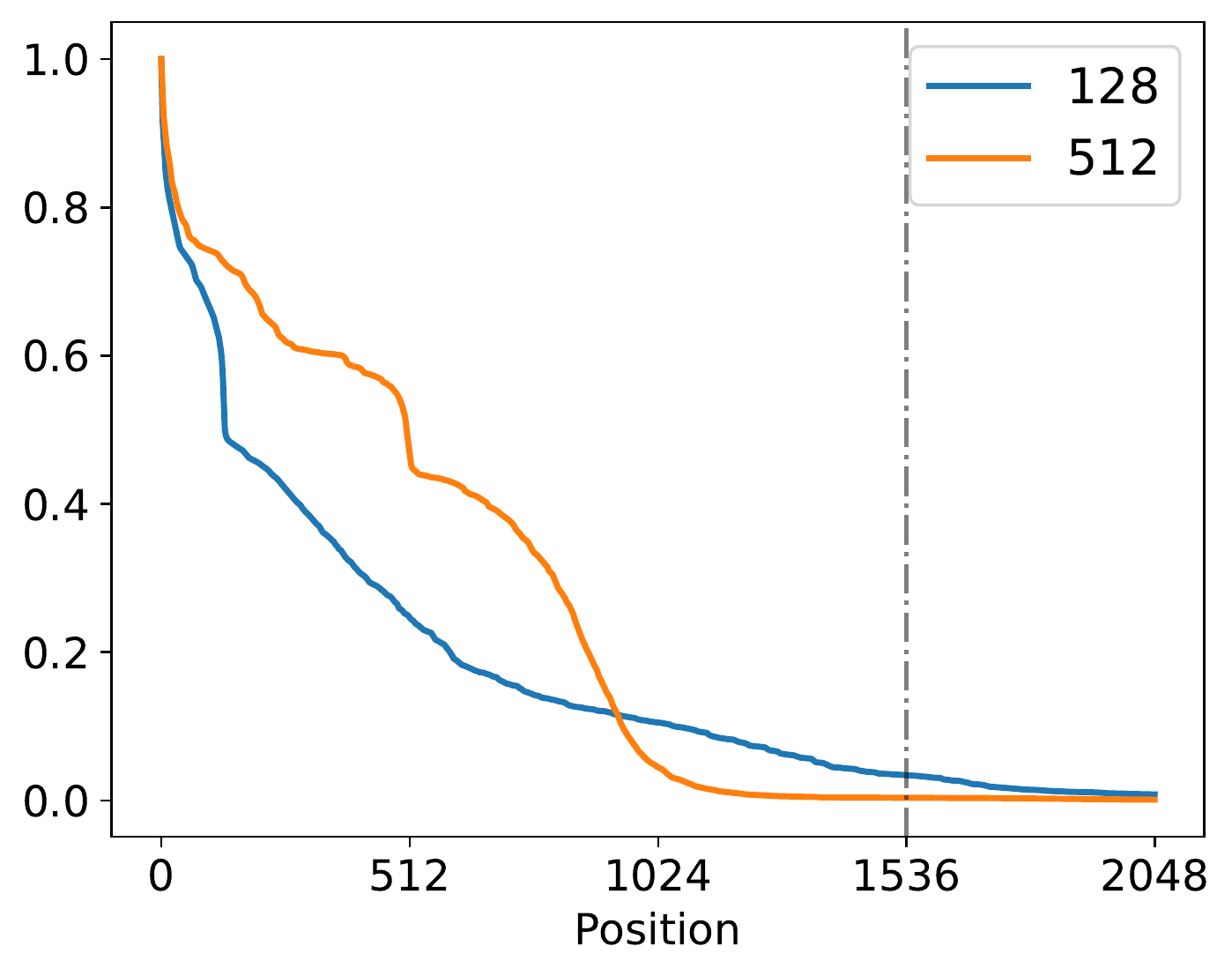}
  \caption{Cumulative normalized gradient 
 of Sinusoidal \\ on ArXiv when predicting the last (2048-th) token\\ with $L_{tr}\in[128, 512]$.}\label{fig:sinusoidal_128_512}
\endminipage
\end{figure*}

\begin{table*}
\begin{align}
    & (\bm W_q(\bm e_m +\bm p_m))^\top (\bm W_k (\bm e_n + \bm p_n)) = \label{eq:decomposition}\\ \nonumber
    &\underbrace{\bm e_m^\top\bm W_q^\top
	        \bm W_k^{\vphantom{\top}}\bm e_n^{\top}}_\text{semantic info.}
         + \underbrace{\bm e_m^\top\bm W_q^\top\bm W_k\bm p_n + \bm p_m^\top\bm W_q^\top\bm W_k\bm e_n + \bm p_m^\top\bm W_q^\top\bm W_k\bm p_n}_\text{mixture of semantic and positional info.} \approx \underbrace{\bm e_m^\top\bm W_q^\top
	        \bm W_k^{\vphantom{\top}}\bm e_n^{\top}}_\text{semantic info.}
         + \underbrace{\bm p_m^\top\bm p_n}_\text{positional info.}
\end{align}
\end{table*}

\subsection{Analyses of Sinusoidal and Rotary}
Sinusoidal~\cite{vaswani2017attention} constructs the positional embedding at position $m$ and $\forall i\in [1,d/2]$ as:
\begin{align}
    \bm p_{m, 2i} =& \sin\left(\frac{m}{10000^{2i/d}}\right), \nonumber \\
    \bm p_{m, 2i+1} =& \cos\left(\frac{m}{10000^{2i/d}}\right)
    \label{eq:sinusoidal}
\end{align}
They will be added with the input embeddings $\{\bm e_m\}_{m=1}^L$ followed by the query and key transformations as shown in Eq.~(\ref{eq:decomposition}). Unlike addition,
Rotary~\cite{su2021roformer} multiplies each token embedding $\bm e_m$ with a position-specific rotation matrix $\bm R_m \bm e_m$.

What could $c_m$ tell us when it is applied to the non-extrapolatable Sinusoidal and Rotary positional embeddings?  As we can see in Figure~\ref{fig:rotary_512} and~\ref{fig:sinusoidal_128_512}, they both fail to focus on the most recent $L_{tr}$ tokens because neither of their formulations guarantees a $L_{tr}$-bounded receptive field. Figure~\ref{fig:sinusoidal_128_512} tells additional stories: To predict the last token (2048-th), Sinusoidal focuses on the 512-th token when $L_{tr}=512$ and the 128-th token when $L_{tr}=128$ as indicated by the sudden jump on their normalized gradient plots. This is because the model has only seen at most $L_{tr}$ positional embeddings and overfitted on them, which provides explicit evidence to the Sinusoidal, or APE in general, overfitting hypothesis made by the author of ALiBi\footnote{\url{https://twitter.com/OfirPress/status/1435690039925567489}}. 
It also explains why RPE is a better choice for length extrapolatable transformers: They cannot overfit on the positional embeddings.

\section{A New RPE for Length Extrapolation}
\subsection{Introduction to Sandwich}
\label{sec:intro_sandwich}
We fix the overfitting issue of Sinusoidal by transforming it into a new RPE, \textbf{Sandwich}, shown in Eq.~(\ref{eq:decomposition}). Specifically, we drop the cross terms and keep only the inner product of two positional embeddings\footnote{We set $\bm p_{m,n}$ to $2d$ as doing so gives better empirical performance; it only needs to be computed once before training.} at $m$ and $n$. Now $\bm p_m^\top \bm p_n$ with $m,n\in[1,L]$ become the temporal bias terms of Sandwich:
\begin{align*}
    \bm p_m^\top \bm p_n = \sum_{i=1}^{\bar d/2} &\sin\left(\frac{m}{10000^{2i/\bar d}}\right) \sin\left(\frac{n}{10000^{2i/\bar d}}\right) + \nonumber \\ &\cos\left(\frac{m}{10000^{2i/\bar d}}\right) \cos\left(\frac{n}{10000^{2i/\bar d}}\right) \nonumber\\
    =\sum_{i=1}^{\bar d/2}&\cos\left(\frac{m-n}{10000^{2i/\bar d}}\right)
\end{align*}
A similar observation was previously made in a context different from length extrapolation~\cite{yan2019tener}.

\begin{figure*}[!htb]
\minipage{0.5\textwidth}
  \includegraphics[width=\linewidth]{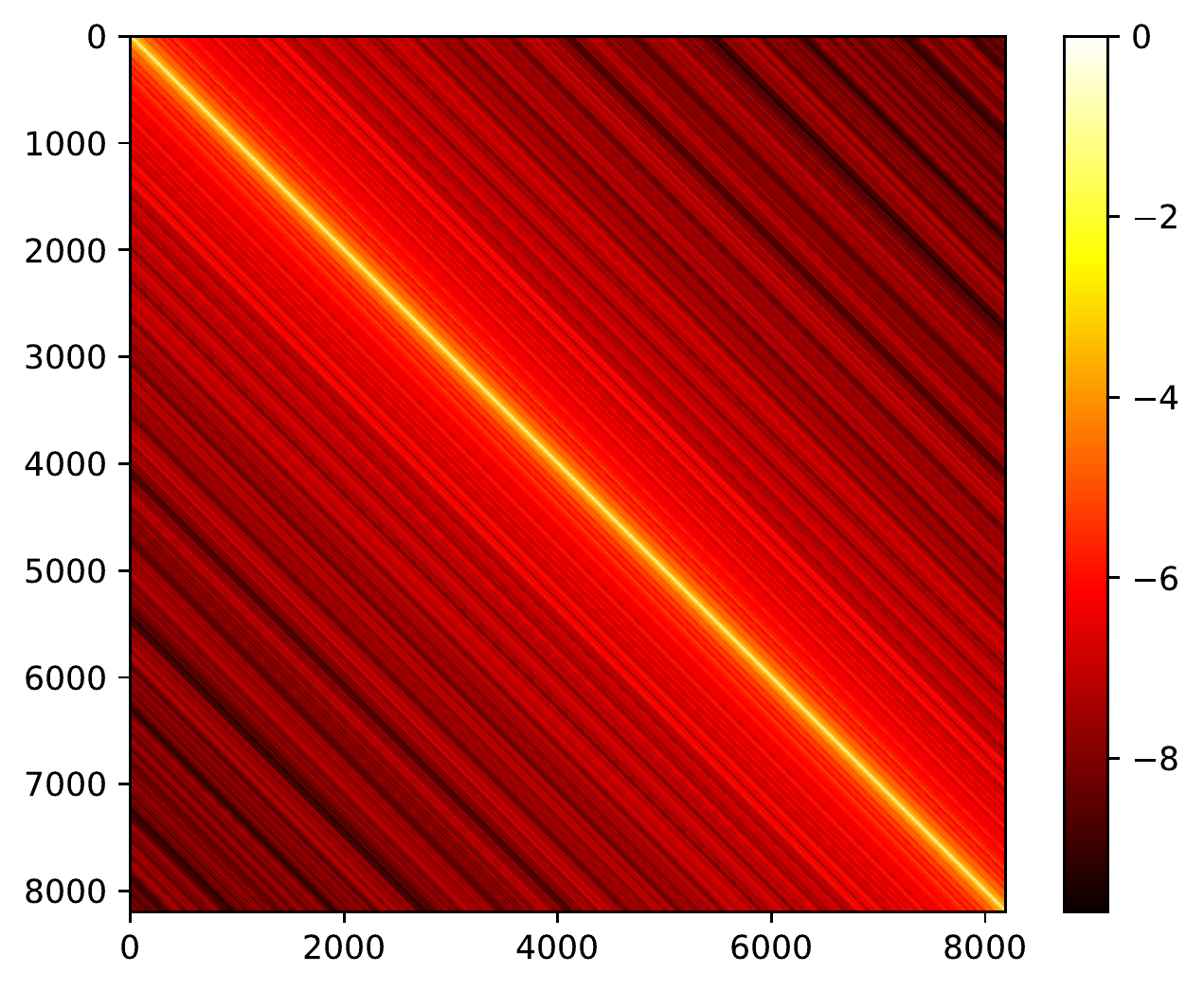}
  \caption{The visualization of Eq.~(\ref{eq:sandwich}) when the \\ compression ratio $h=8$ and $\bar d=128$.}\label{fig:heatmap_8}
\endminipage\hfill
\minipage{0.5\textwidth}
  \includegraphics[width=\linewidth]{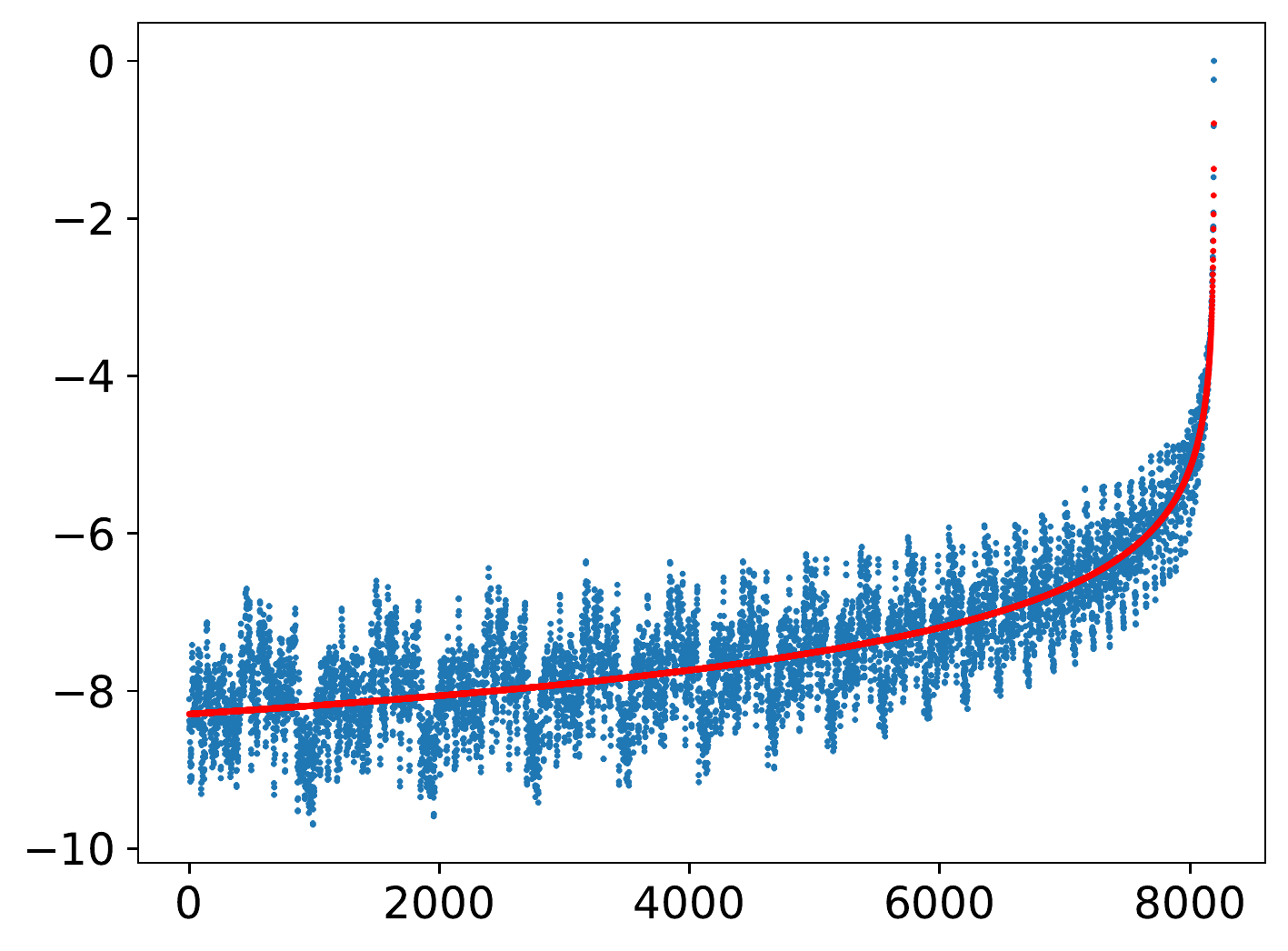}
  \caption{We plot the last row in Figure~\ref{fig:heatmap_8}. The red curve is the least-squared fitted log function: $y = -0.825\cdot\log(|m-n|)+1)-0.8$ with $m=8192$ in this example.}\label{fig:diagonal_8}
\endminipage\hfill
\end{figure*}

\begin{figure}[!htb]
  \includegraphics[width=\linewidth]{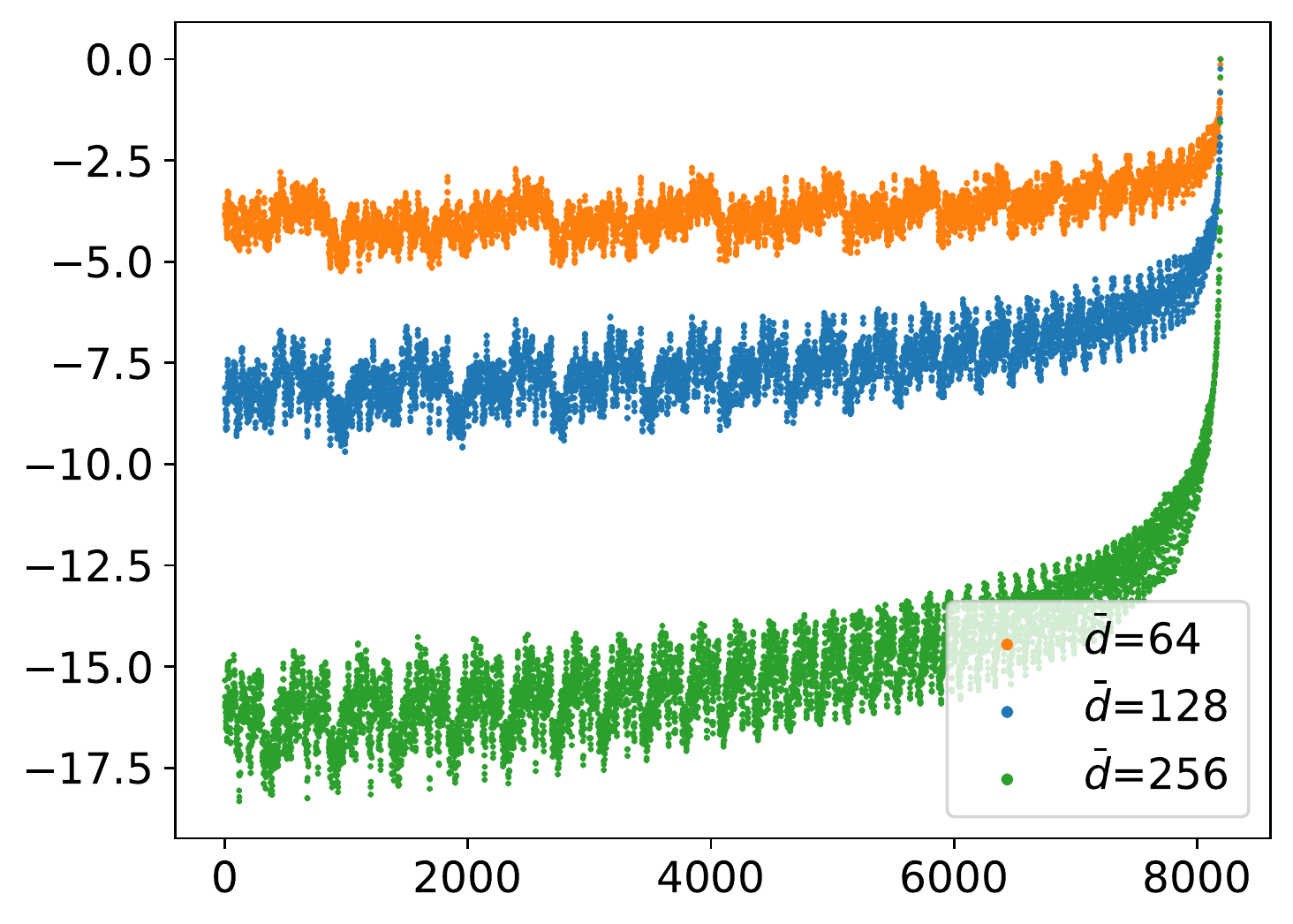}
  \caption{We experiment with different $\bar d$ and find they create different windowed attention effect.}\label{fig:different_d}
\end{figure}

The largest value of $\bm p_m^\top \bm p_n$ happens at the point where $m-n=0$, which gives the maximum value of $\bar d / 2$. 
To align $L_{tr}$ with the ERF of Sandwich, we need to further check that $\bm p_m^\top \bm p_n$ demonstrates a similar windowed attention effect as ALiBi. 
This can be done by subtracting all $\bm p_m^\top \bm p_n$ by $\bar d / 2$ and further dividing them by a set of predefined compression ratios. 
for the sake of simplicity, we set the compression ratios to be the same as ALiBi's $h=n\cdot\frac{8}{H}$ with $n\in\{1\dots H\}$:
\begin{equation}
    \frac{\bm p_m^\top \bm p_n - \bar d/2}{h}
    \label{eq:sandwich}
\end{equation}
Eq.~(\ref{eq:sandwich}) is added after the scaled softmax is done in Eq.~(\ref{eq:scaled_softmax}).
Figures~\ref{fig:heatmap_8} and~\ref{fig:diagonal_8} show a visualization of Sandwich when $h=8$.
Sandwich indeed has the same decay-with-distance pattern as ALiBi.\footnote{Fun fact: We imagine different compression ratios as the ways we eat sandwiches: For a huge sandwich, we have to squeeze it more to fit in our mouths!}

Note that we deliberately decouple this $\bar d$ from $d$ in Eq.~(\ref{eq:sinusoidal}) since we treat $\bar d$ as a hyperparameter that controls the shape of Sandwich. A larger $\bar d$ leads to a stronger windowed attention effect as shown in Figure~\ref{fig:different_d}. We set $\bar d=128$ in this work for all the experiments. We also experiment with smaller and larger $\bar d$ and only find worse performance. Finally, readers can find the reference Python implementation in Appendix~\ref{sec:python_impl}.

\newcommand{\rowresults}[9]{ \centering #1 & #2 & #8 & #4 & #7 & #6 & #9 & #3 & #5 \\}

\begin{table*}[!ht]
    \setlength{\tabcolsep}{2pt}
    \hspace{-3.5mm}
    \begin{tabular}{@{\extracolsep{3pt}}lcclcccll}
    \hline\hline
    \multicolumn{9}{c}{\textbf{OpenWebText2}}\\
    \cline{1-6} \cline{8-9}
    \rowresults{$L_{ex}$}{\multicolumn{1}{c}{Sandwich}}{\multicolumn{1}{c}{KERPLE}}{\multicolumn{1}{c}{ALiBi}}{\multicolumn{1}{c}{T5}}{Rotary}{Sinusoidal}{\multicolumn{1}{c}{Smoothed}}{\quad}
    \rowresults{512}{23.5 $\pm$ 3.8}{\textbf{22.6 $\pmb{\pm}$ 3.5}$^\ast$}{\textbf{22.8 $\pmb{\pm}$ 3.3}}{22.6 $\pm$ 3.6$^\ast$}{23.0 $\pm$ 3.4$^\ast$}{26 $\pm$ 1$^\dagger$}{23.2 $\pm$ 3.7}{}
    \rowresults{1024}{\textbf{23.0 $\pmb{\pm}$ 3.6}}{\textbf{22.0 $\pmb{\pm}$ 3.3}$^\ast$}{23.3 $\pm$ 3.4}{22.2 $\pm$ 3.3$^\ast$}{61$^\dagger$}{14168$^\dagger$}{23.1 $\pm$ 3.6}{}
    \rowresults{2048}{23.3 $\pm$ 3.5}{\textbf{21.9 $\pmb{\pm}$ 3.1}$^\ast$}{23.5 $\pm$ 3.3}{23.0 $\pm$ 3.1}{96$^\dagger$}{20370$^\dagger$}{\textbf{23.2 $\pmb{\pm}$ 3.2}}{}
    \rowresults{4096}{23.8 $\pm$ 3.3}{\textbf{22.1 $\pmb{\pm}$ 2.9}$^\ast$}{\textbf{23.5 $\pmb{\pm}$ 3.3}$^\ast$}{26.8 $\pm$ 3.2$^\dagger$}{232$^\dagger$}{42003$^\dagger$}{23.6 $\pm$ 3.0}{}
    \rowresults{8192}{24.7 $\pm$ 3.4}{\textbf{22.3 $\pmb{\pm}$ 2.9}$^\ast$}{\textbf{23.5 $\pmb{\pm}$ 3.3}$^\ast$}{38.6 $\pm$ 7.2$^\dagger$}{343$^\dagger$}{67869$^\dagger$}{24.0 $\pm$ 2.9}{}
    \hline\hline
    \multicolumn{9}{c}{\textbf{ArXiv}}\\
    \cline{1-6} \cline{8-9}
    \rowresults{$L_{ex}$}{\multicolumn{1}{c}{Sandwich}}{\multicolumn{1}{c}{KERPLE}}{\multicolumn{1}{c}{ALiBi}}{\multicolumn{1}{c}{T5}}{Rotary}{Sinusoidal}{\multicolumn{1}{c}{Smoothed}}{\quad}
    \rowresults{512}{5.27 $\pm$ 0.33}{5.22 $\pm$ 0.37}{\textbf{5.25 $\pmb{\pm}$ 0.33}}{\textbf{5.16 $\pmb{\pm}$ 0.37}$^\ast$}{\textbf{5.25 $\pmb{\pm}$ 0.33}}{5.8$^\dagger$}{5.33 $\pm$ 0.32}{}
    \rowresults{1024}{\textbf{5.05 $\pmb{\pm}$ 0.33}}{4.95 $\pm$ 0.34$^\ast$}{5.41 $\pm$ 0.36$^\dagger$}{\textbf{4.91 $\pmb{\pm}$ 0.35}$^\ast$}{16.02$^\dagger$}{1070$^\dagger$}{5.13 $\pm$ 0.32}{}
    \rowresults{2048}{\textbf{5.02 $\pmb{\pm}$ 0.34}}{\textbf{4.83 $\pmb{\pm}$ 0.35}$^\ast$}{5.58 $\pm$ 0.40$^\dagger$}{4.92 $\pm$ 0.35$^\ast$}{33.76$^\dagger$}{1784$^\dagger$}{5.15 $\pm$ 0.36}{}
    \rowresults{4096}{\textbf{5.15 $\pmb{\pm}$ 0.39}}{\textbf{4.84 $\pmb{\pm}$ 0.34}$^\ast$}{5.58 $\pm$ 0.40$^\dagger$}{5.35 $\pm$ 0.36}{71.96$^\dagger$}{18050$^\dagger$}{5.33 $\pm$ 0.39}{}
    \rowresults{8192}{\textbf{5.28 $\pmb{\pm}$ 0.44}}{\textbf{4.90 $\pmb{\pm}$ 0.33}$^\ast$}{5.58 $\pm$ 0.40$^\dagger$}{6.74 $\pm$ 0.90$^\dagger$}{111$^\dagger$}{44100$^\dagger$}{5.45 $\pm$ 0.42}{}
    \hline\hline
    \multicolumn{9}{c}{\textbf{GitHub}}\\
    \cline{1-6} \cline{8-9}
    \rowresults{$L_{ex}$}{\multicolumn{1}{c}{Sandwich}}{\multicolumn{1}{c}{KERPLE}}{\multicolumn{1}{c}{ALiBi}}{\multicolumn{1}{c}{T5}}{Rotary}{Sinusoidal}{\multicolumn{1}{c}{Smoothed}}{\quad}
    \rowresults{512}{2.88 $\pm$ 0.12}{2.81 $\pm$ 0.14$^\ast$}{2.83 $\pm$ 0.11$^\dagger$}{\textbf{2.76 $\pmb{\pm}$ 0.14}$^\ast$}{\textbf{2.82 $\pmb{\pm}$ 0.11}}{4$^\dagger$}{2.88 $\pm$ 0.17}{}
    \rowresults{1024}{2.71 $\pm$ 0.09}{2.67 $\pm$ 0.10$^\ast$}{2.97 $\pm$ 0.11$^\dagger$}{\textbf{2.61 $\pmb{\pm}$ 0.08}$^\ast$}{3.86 $\pm$ 0.25$^\dagger$}{8342$^\dagger$}{\textbf{2.70 $\pmb{\pm}$ 0.07}}{}
    \rowresults{2048}{\textbf{2.69 $\pmb{\pm}$ 0.11}}{2.65 $\pm$ 0.10$^\ast$}{3.01 $\pm$ 0.10$^\dagger$}{\textbf{2.65 $\pmb{\pm}$ 0.05}}{5.94 $\pm$ 0.64$^\dagger$}{9179$^\dagger$}{2.74 $\pm$ 0.08}{}
    \rowresults{4096}{\textbf{2.73 $\pmb{\pm}$ 0.12}}{\textbf{2.70 $\pmb{\pm}$ 0.09}}{3.01 $\pm$ 0.10$^\dagger$}{2.91 $\pm$ 0.12}{11.1 $\pm$ 1.55$^\dagger$}{11017$^\dagger$}{2.78 $\pm$ 0.08}{}
    \rowresults{8192}{\textbf{2.79 $\pmb{\pm}$ 0.15}}{\textbf{2.75 $\pmb{\pm}$ 0.08}}{3.01 $\pm$ 0.10$^\dagger$}{3.68 $\pm$ 0.50$^\dagger$}{20.2 $\pm$ 2.75$^\dagger$}{11270$^\dagger$}{2.83 $\pm$ 0.08}{}
    \hline\hline
    \end{tabular}
    \caption{\textbf{Perplexity Comparison on the OpenWebText2, GitHub, and ArXiv datasets.} All models are trained for 50k steps with a training length of 512 and five random seeds. The models in the left section have parameter-free positional embeddings. In contrast, both KERPLE and T5 are equipped with learnable parameters. A fair comparison should only be made within the same section. $x^\dagger$ means sandwich is statistically significantly~\emph{better} than $x$. $x^\ast$ means sandwich is statistically significantly~\emph{worse} than $x$. The test used is paired two-sided t-test with $\alpha=0.05$. More details about the datasets and hyperparameters are provided in Appendix~\ref{sec:artifact} and~\ref{sec:implementation_details}.}
    \label{tab:openweb-github-arxiv}
\end{table*}

\subsection{Experiments and Discussion}
To verify the performance of Sandwich, we train a transformer language model following previous work~\cite{press2022train,chi2022kerple}. Table~\ref{tab:openweb-github-arxiv} presents the results; the left part contains all models without learnable parameters, and the right part contains models with learnable parameters. These numbers should not be compared across sections. 

In general, models on the right achieve lower perplexities across the three datasets. This is expected as they can adapt to individual datasets more easily thanks to the additional learnable parameters. 
However, there is no free lunch: They often consume more GPU memory and run much slower. For example, T5 is 10\% slower than Sandwich during the training stage. Note that Sandwich can also be equipped with learnable parameters such as learnable compression ratios $h$; this is left to future work.
We now shift our focus to the left section. When $L_{ex}=L_{tr}=512$, Sandwich is comparable to other models except that Rotary performs a bit better on  OpenWebText2. 
Once we increase $L_{ex}$, Sandwich begins to reveal its advantages: On ArXiv and GitHub, it is consistently better than all the baselines but only marginally worse than ALiBi when $L_{ex}\geq 4096$ on OpenWebText2.

\begin{figure}[!htb]
  \includegraphics[width=\linewidth]{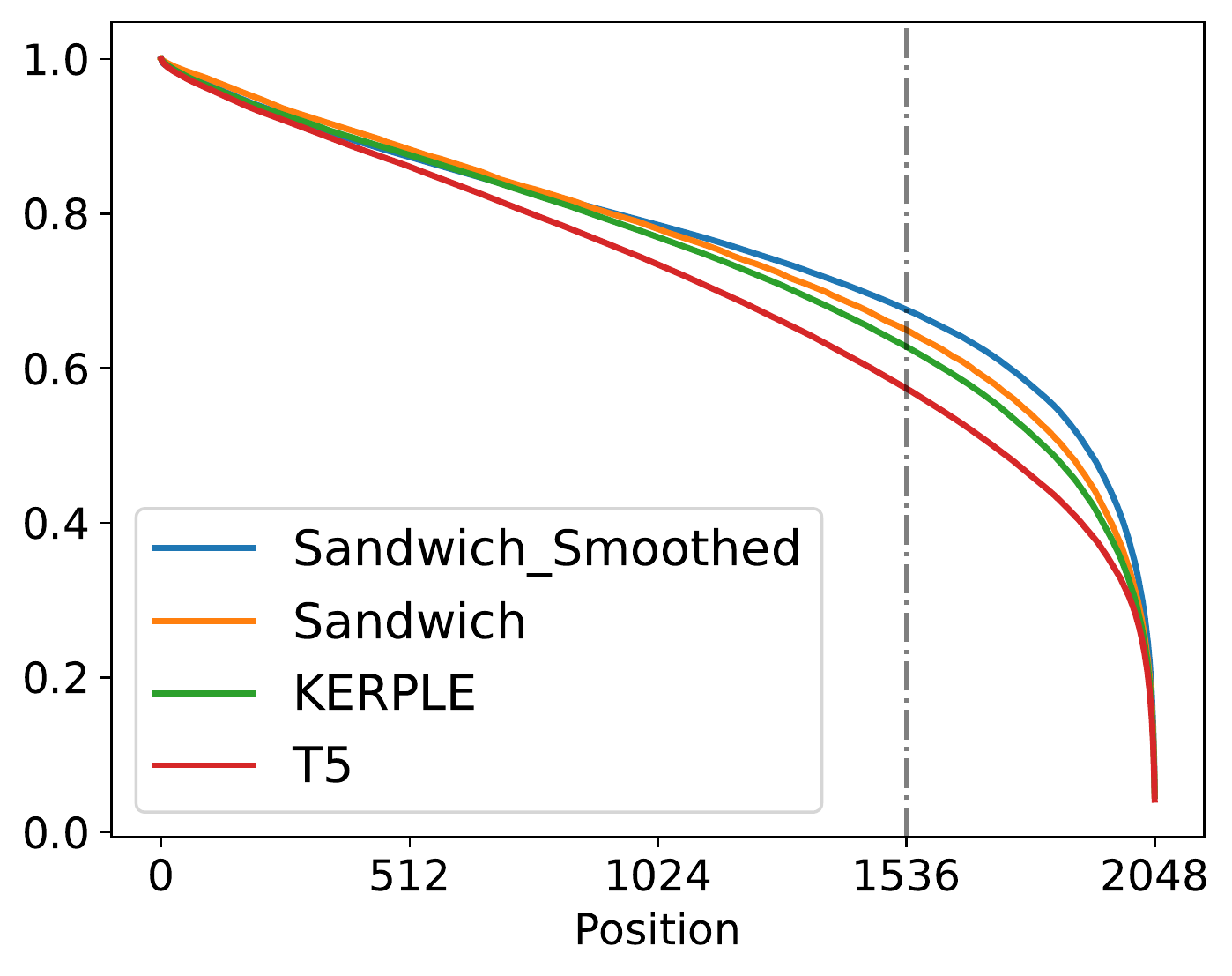}
  \caption{Cumulative normalized gradient of Sandwich, Smoothed Sandwich, KERPLE, and T5 on ArXiv when predicting the last (2048-th) token with $L_{tr}=512$.}\label{fig:sandwich_kerple_t5_receptive}
\end{figure}

It is worth mentioning that Sandwich is the first parameter-free RPE that truly makes use of distant token information beyond $L_{tr}=512$. To see this, notice that lower (better) perplexities occur at $L_{ex}>L_{tr}=512$.
The gradient analysis tool in \S\ref{sec:quant_cm} further corroborates this in Figure~\ref{fig:sandwich_kerple_t5_receptive}, which reveals a  receptive field pattern distinct from that of ALiBi and windowed attention. Even though Sandwich allocates about 60\% of the total cumulative gradient on the most recent $L_{tr}=512$ tokens, distant tokens beyond $L_{tr}$ still contribute substantially to the model prediction.

\emph{Why do ALiBi and windowed attention need to have their ERFs covered by $L_{tr}$ while Sandwich does not?} To answer this question, we revisit Figure~\ref{fig:diagonal_8} and approximate (least-squared) the original temporal bias pattern using a log curve, which gives a snug fit\footnote{In the actual implementation, we fit the curve using the most recent 50 points of Sandwich. The reason is because the most recent tokens are more important, and we want them to be closer to the original Sandwich.}: $y = -0.825\cdot\log\left(1+|m-n|\right)-0.8$. Table~\ref{tab:openweb-github-arxiv}  shows its language modeling performance under the ``smoothed'' column. 
Pictorially, the log curve decays relatively fast when two tokens are nearby and plateaus when the distance between them increases. In other words, tokens that are far away from the last one ($m=8192$) share similar temporal biases, possibly leading to beneficial averaging and denoising effects.
Note that the averaging effect does not come out of thin air during the extrapolation stage: The almost linear segment ranging from 1536 to 1792 suggests that Sandwich was trained to perform averaging within $L_{tr}$; it just needs to average over more historical tokens when it extrapolates to longer $L_{ex}$.
In contrast, ALiBi's linear bias lacks the middle ground to learn the averaging behavior: It either decays so fast that distant tokens are masked out or so slow that the ERF becomes much greater than $L_{tr}$.
The averaging hypothesis also explains why Sandwich, KERPLE, and T5's perplexities go up in Table~\ref{tab:openweb-github-arxiv} instead of continuing to decrease after some $L_{ex}$ (4096 on ArXiv for example): While averaging and denoising improve performance, doing so over too many historical tokens (very large $L_{ex}$) will reintroduce noises.

\subsection{Connection to KERPLE and T5}
KERPLE~\cite{chi2022kerple} has the formulation of $c -r_1\cdot\log\left(1+r_2|m-n|\right)$. The $-0.8$ in our fitted log curve term can be absorbed by $c$, as Softmax is shift-invariant, and if we set $r_1=0.825$ and $r_2=1$, Sandwich becomes a special case of KERPLE.
T5~\cite{raffel2019exploring} adopts the log-binning strategy that assigns distinct bins to nearby tokens whereas distant tokens all share the same bin. In spirit, T5 treats distant tokens similarly to Sandwich.
Figure~\ref{fig:sandwich_kerple_t5_receptive} verifies that all three of them share a similar empirical receptive field pattern.

\section{Conclusion}
In this paper, we first establish the connection between ALiBi and windowed attention through their constructions and language modeling performance. We then develop a cumulative normalized gradient tool to measure the empirical receptive field. It shows that length extrapolation of ALiBi and windowed attention is possible when the training sequence length covers the empirical receptive field. It also reveals the models' limitation of not utilizing information beyond the training sequence length.
Fortunately, this is overcome by our new relative positional embedding, Sandwich, which is simplified from the earliest proposed Sinusoidal positional embedding.
Finally, Sandwich demonstrates a  log-decaying temporal bias pattern similar to that previously seen in the design of KERPLE and T5, and such pattern is likely to be the secret to successful length extrapolation.
Together these findings supports more effective design of future extrapolatable transformer language models.

\section*{Limitations}
Although Sandwich, KERPLE, and T5 use information beyond training sequence length, their receptive fields still highly favor the most recent tokens. While this recency bias is beneficial to the modeling of human-written text, it is problematic in other scenarios. 

Let us consider the task of~\emph{parity} prediction: A model needs to predict whether a bit string has an even or odd number of ones. For example, the parity of [1, 1, 0, 1] is odd (or 1) and the parity of [1, 0, 1, 0] is even (or 0). Unlike human-written text, every single bit is equally important. Transformer language models with current RPEs still struggle on this simple task~\cite{anil2022exploring}. Its difficulty can be explained by the recency bias effect that we described.
Devising a new positional embedding or transformer model architecture that solves this problem is a promising direction for future work.

\section*{Ethics Statement}
Our work advances the understanding of positional embeddings adopted in almost all transformer models. In addition, our proposed new positional embedding significantly reduces energy consumption and training cost thanks to its length extrapolation property.
Finally, our work lays the groundwork for developing future transformers that are greener and more cost-efficient enabled by improved length extrapolation.
Inappropriate usage of our technique might have negative societal impacts. These include the ethical challenges of improper text generation and  privacy issues inherent in the data collection process. 
These implications apply to any natural language processing research and are not unique to this specific work.

\section*{Acknowledgment}
The authors acknowledge the support from
Boeing (2019-STU-PA-259), 
Amazon (CC ADV 00474341 2021 TR),  
NSF MRI Award 1919452, and
Princeton Research Computing.

\bibliography{custom}
\bibliographystyle{acl_natbib}
\clearpage
\appendix

\section{Results on OpenWebText2}
\begin{table*}[!ht]
    \setlength{\tabcolsep}{2pt}
    \hspace{-7mm}
    \begin{tabular}{@{\extracolsep{3pt}}lccccccccccccccccccccc}
    \hline\hline
    \multirow{2}{*}{$L_{ex}$} & \multicolumn{6}{c}{Shift all $h$ by $\Delta$} & & \multicolumn{5}{c}{Same $h$ for all heads} & & \multicolumn{6}{c}{Windowed Attention Size $w$} \\
     \cline{2-7} \cline{9-13} \cline{15-20} &
     $\Delta$:-3 & 0 & 2 & 4 & 6 & 8 & & $h$:0 & 2 & 4 & 6 & 8 & & $w$:40 & 80 & 100& 120 & 160 & 320 \\
    \hline
    512 & 18.6 & 19.0 & 19.5 & 20.0 & 20.5 & 20.5 & & 32.7 & 22.2 & 19.7 & 19.7 & 20.5 & & 25.3 & 23.7 & 23.1 & 24.0 & 22.9 & 21.9 \\
    1024 & 21.6 & 19.3 & 19.6 & 24.8 & 232 & 232 & & 32.8 & 23.2 & 24.9 & 146 & 232 & & 25.3 & 23.7 & 23.2 & 137 & 234 & 353 \\
    2048 & 21.6 & 19.7 & 20.5 & 29.3 & 299 & 299 & & 32.8 & 23.2 & 24.9 & 165 & 299 & & 25.3 & 23.7 & 23.2 & 137 & 236 & 408 \\
    4096 & 21.6 & 19.7 & 20.5 & 29.4 & 299 & 299 & & 32.9 & 23.2 & 24.9 & 165 & 299 & & 25.3 & 23.7 & 23.2 & 137 & 236 & 408 \\
    8192 & 21.6 & 19.7 & 20.5 & 29.4 & 299 & 299 & & 32.9 & 23.2 & 24.9 & 165 & 299 & & 25.3 & 23.7 & 23.2 & 137 & 236 & 408 \\
    \hline\hline
    \end{tabular}
    \caption{The three experiments on the OpenWebText2 dataset.}
    \label{tab:openwebtext2}
\end{table*}

\begin{figure*}[]
\minipage{0.5\textwidth}
  \includegraphics[width=\linewidth]{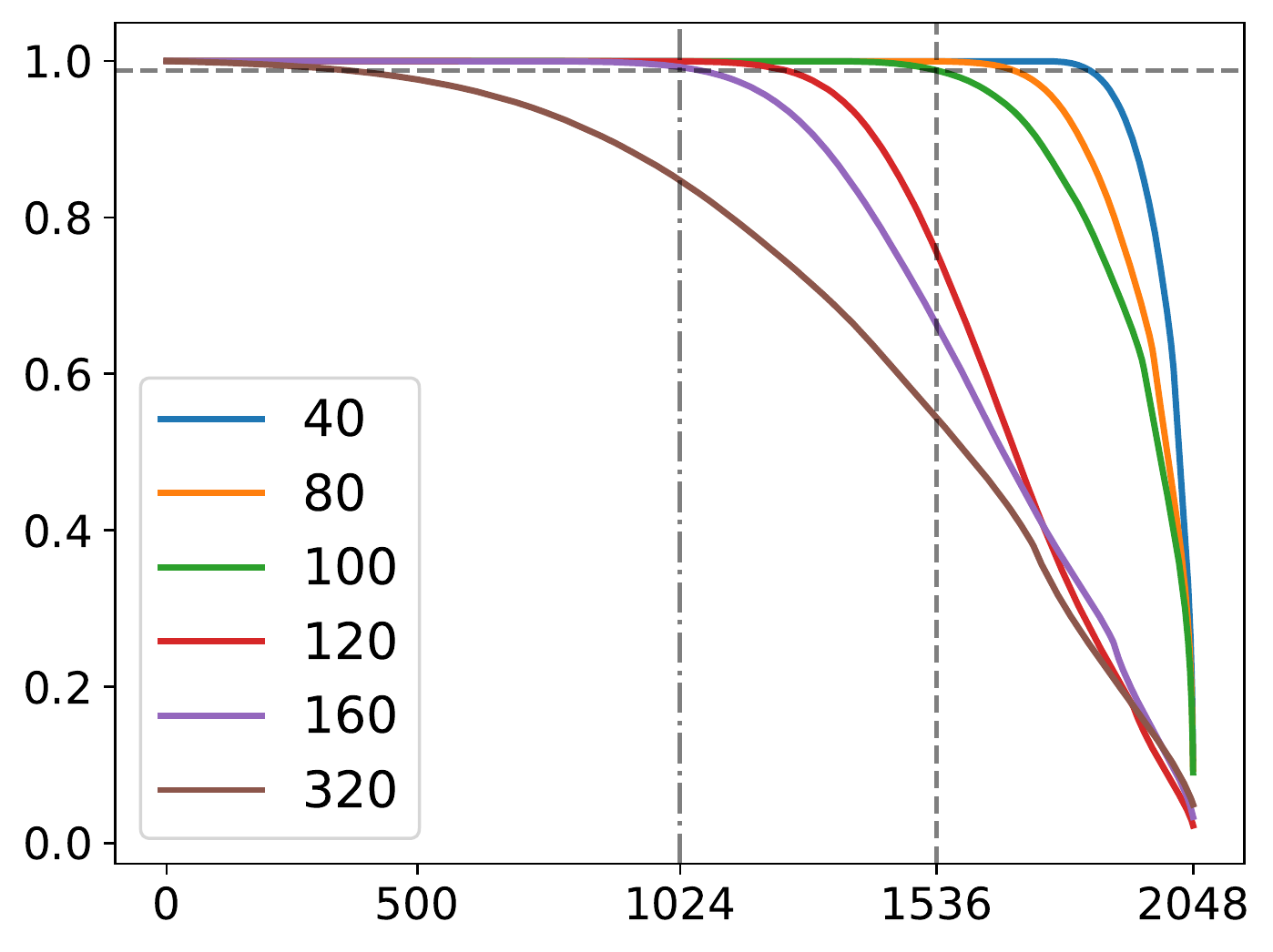}
  \caption{Cumulative normalized gradient 
 on \\ OpenWebText2 when predicting the last (2048-th) \\ token. Windowed Attention Size $w=$}\label{fig:openwebtext2_sec3}
\endminipage\hfill
\minipage{0.5\textwidth}
\includegraphics[width=\linewidth]{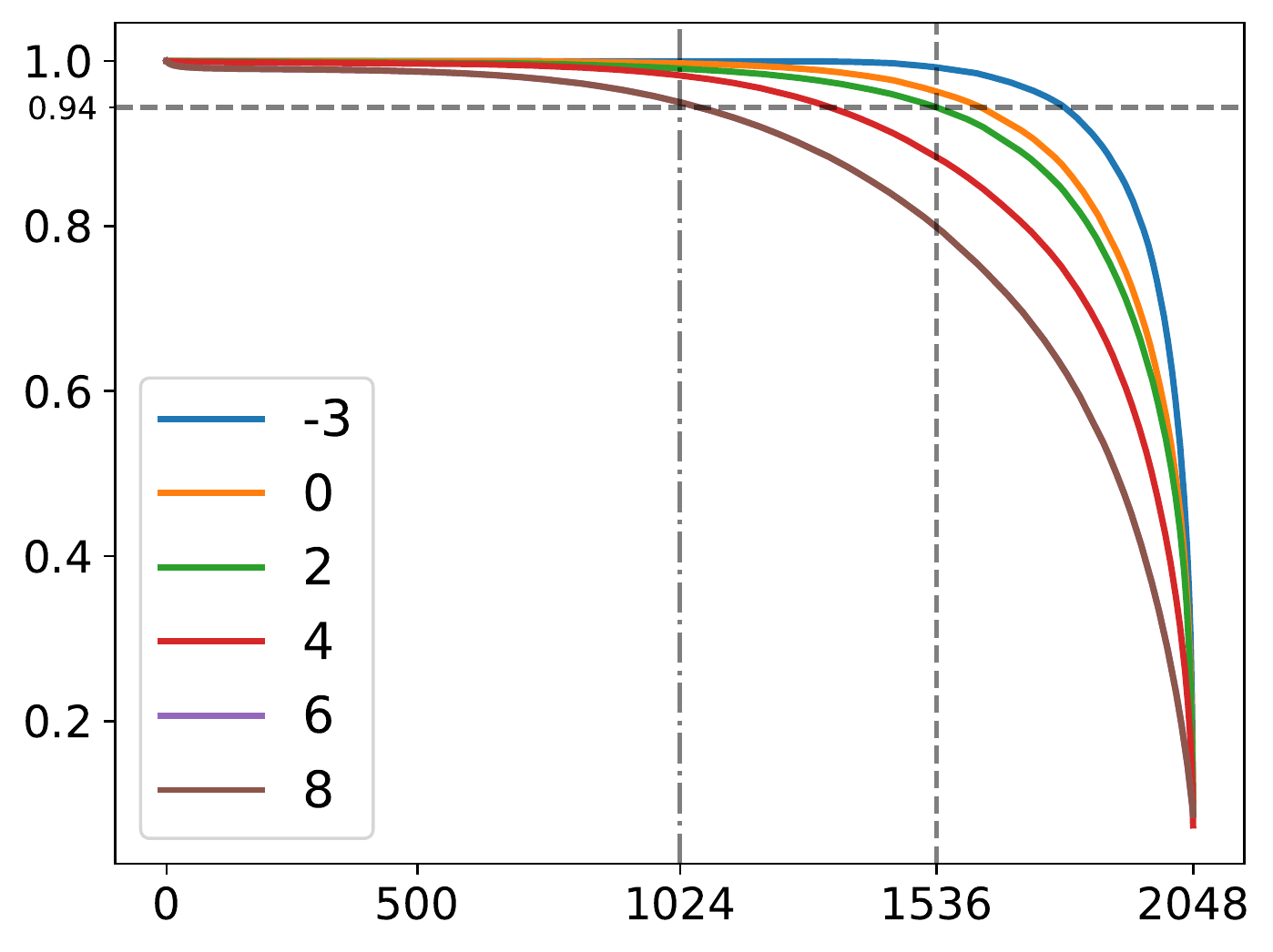}
\caption{Cumulative normalized gradient 
 on \\ OpenWebText2 when predicting the last (2048-th) token. Shift all $h$ by $\Delta=$}\label{fig:openwebtext2_sec1}
\endminipage
\end{figure*}

Table~\ref{tab:openwebtext2} includes the three experiments conducted in \S\ref{sec:three_exps} on OpenWebText2. Their corresponding receptive field plots are shown in Figure~\ref{fig:openwebtext2_sec3} and~\ref{fig:openwebtext2_sec1}.
\label{sec:appendix}

\section{Efficient Inference}
\label{sec:efficient_inference}
Although ALiBi might not be using token information further than $L_{tr}$, it has the nice property of efficient inference~\cite{usecase}. 
Tables~\ref{tab:arxiv} and~\ref{tab:openwebtext2} show that ALiBi perplexities stay constant when $L_{ex}\geq 2048$. This suggests a cache window size $\bar w=2048$ for inference. 
The generation of the first $\bar w$ tokens remains the same, and we can still cache all  $\bm q_m$, $\bm k_m$, and $\bm v_m$ vectors for $m\in[1,2048]$. 
When it comes to generating the $\bar w + 1$-th token, we simply discard the first cached $\bm q_1$, $\bm k_1$, and $\bm v_1$ and use the rest of $\bar w - 1$ tokens along with the newly added token to perform self-attention.
If we want to generate a length $L_{ex}$ text snippet, the complexity is $O(\bar w\times L_{ex})$ instead of $O(L_{ex}^2)$. 
This complexity is also better than that of an APE model, which is $O(\bar w^2\times L_{ex})$ since an APE model needs to completely re-encode the previous $\bar w$ vectors when generating  new tokens following the first $\bar w$ ones.

We implement the process discussed above to verify that ALiBi indeed allows for efficient inference.  The results, along with ones for Sandwich, are presented in Table~\ref{tab:efficient_inference}. 
Both ALiBi and Sandwich permit efficient inference by setting $\bar w=2048$. It is worth pointing out that the performance of Sandwich at $L_{ex}=4096$ becomes a bit worse compared to that in Table~\ref{tab:openweb-github-arxiv}. This is more evidence that Sandwich is using longer than $L_{tr}$ token information.

\begin{table}[!htbp]
    \setlength{\tabcolsep}{2pt}
    {
    \begin{tabular}{@{}lcccccccc}
    \hline\hline
    \multirow{2}{*}{$L_{ex}$} & \multicolumn{2}{c}{OpenWebText2} & & \multicolumn{2}{c}{Arxiv} & & \multicolumn{2}{c}{GitHub}\\
     \cline{2-3} \cline{5-6} \cline{8-9} &
     Sandwich & ALiBi & & Sandwich & ALiBi & & Sandwich & ALiBi\\
    \hline
    4096 & 23.9 & 23.5 & & 5.31 & 5.59 & & 2.79 & 3.01\\
    8192 & 24.1 & 23.5 & & 5.35 & 5.59 & & 2.81 & 3.01\\
    16384 & 24.1 & 23.5 & & 5.35 & 5.59 & & 2.81 & 3.01\\
    \hline\hline
    \end{tabular}
    }
    \caption{Efficient Inference with $\bar w=2048$.}
    \label{tab:efficient_inference}
\end{table}

\section{Scientific Artifacts}
\label{sec:artifact}
\begin{table}[!ht]
    \centering
    \begin{tabular}{lccc}
    \hline\hline
    & OpenWebText2 & GitHub & ArXiv\\ \hline
    Raw Size & 66.77 GB & 95.16 GB & 56.21 GB\\
    Type & Internet & Coding & Academic\\
    \hline\hline
    \end{tabular}
    \caption{\textbf{Dataset Overview.} Raw Size is the size before any up- or down-sampling.}
    \label{tab:dataset}
\end{table}
We use the gpt-neox library~\cite{gpt-neox-library} under Apache-2.0 license and the datasets~\cite{pile} released by the authors of gpt-neox. The codebase and datasets (Table~\ref{tab:dataset}) are publicly released for research purposes. The steps taken to protect the privacy and anonymization are discussed in~\citet{pile} section 6 and 7. Finally,~\citet{pile} section 5 also discusses the distribution and statistics of the datasets used in this work.

\section{Implementation Details}
\label{sec:implementation_details}
The configurations and hyperparameters are outlined in Table~\ref{tab:model_configs}. The pretraining takes 5 hours on a single NVIDIA A-100 GPU. We do not tune any hyperparameters and just use the default ones.
\begin{table}[!ht]
    \centering
    \setlength{\tabcolsep}{3pt}
    \begin{tabular}{ccccc}
        \hline\hline
         \# Layers & Hidden Size & \# Attention Heads & Train Seq. Len. & \# Trainable Params.\\
         12 & 64 & 12 & 512 & ~162M\\ \hline
         Optimizer & Batch Size & Train Steps & Precision & \# Trainable Params. for RPEs\\
         Adam (lr 6e-4) & 32 & 50,000 & bfloat16 & 0\\
         \hline\hline
    \end{tabular}
    \caption{162M Model Configurations.}
    \label{tab:model_configs}
\end{table}

\clearpage
\section{Python Implementation of Sandwich}
\label{sec:python_impl}
\begin{minted}{python}
import numpy as np

base = 1e4
heads = 12
seq_len = 8192
positions = np.arange(seq_len)[..., None]
bar_d = 128 # This is the hyperparameter of Sandwich
i = np.arange(bar_d // 2)

pos_embs = np.concatenate([np.sin(positions / base ** (2 * i / bar_d)),
                          np.cos(positions / base ** (2 * i / bar_d))],
                          axis=-1)
sandwich = np.matmul(pos_embs, pos_embs.T)
compression_ratio = np.arange(1, heads + 1) * 8 / heads
multi_head_sandwich = sandwich[None, ...] / compression_ratio[..., None, None]
\end{minted}

\end{document}